\documentclass[sigconf]{acmart}

\usepackage{algorithm}
\usepackage{algpseudocode}
\usepackage{amsmath}
\usepackage{multirow} 
\usepackage{booktabs}

\makeatletter
\newtheoremstyle{acmrestate}
  {.5\baselineskip\@plus.2\baselineskip\@minus.2\baselineskip}
  {.5\baselineskip\@plus.2\baselineskip\@minus.2\baselineskip}
  {\@acmplainbodyfont}
  {\@acmplainindent}
  {\@acmplainheadfont}
  {.}
  {.5em}
  {\thmname{#1} \thmnote{#3}} 
\theoremstyle{acmrestate}
\newtheorem*{restate_inner}{Theorem} 

\newenvironment{restate}[1]{
  \begin{restate_inner}[\ref{#1}]
}{
  \end{restate_inner}
}
\makeatother

\author{Lian Shen}
\affiliation{%
  \institution{Xiamen University}
  \city{Xiamen}
  \country{China}}
\email{shenlian@stu.xmu.edu.cn}

\author{Zhendan Chen}
\affiliation{%
  \institution{Xiamen University}
  \city{Xiamen}
  \country{China}}
\email{chenzhendan@stu.xmu.edu.cn}

\author{Meijia Song}
\affiliation{%
  \institution{Xiamen University}
  \city{Xiamen}
  \country{China}}
\email{songmj@stu.xmu.edu.cn}

\author{Yinghui jiang}
\affiliation{%
  \institution{Xiamen University}
  \city{Xiamen}
  \country{China}}
\email{stardjyeah@gmail.com}

\author{Ziming Su}
\affiliation{%
  \institution{Xiamen University}
  \city{Xiamen}
  \country{China}}
\email{suziming@stu.xmu.edu.cn}
\author{Juan Liu}
\affiliation{%
  \institution{Xiamen University}
  \city{Xiamen}
  \country{China}}
 \email{cecyliu@xmu.edu.cn} 
\author{Xiangrong Liu}
\affiliation{%
  \institution{Xiamen University}
  \city{Xiamen}
  \country{China}}
\email{xrliu@xmu.edu.cn}

\begin{document}

\title{Decoupling and Damping: Structurally-Regularized Gradient Matching for Multimodal Graph Condensation}


\begin{abstract}
In multimodal graph learning, graph structures that integrate information from multiple sources, such as vision and text, can more comprehensively model complex entity relationships. However, the continuous growth of their data scale poses a significant computational bottleneck for training. Graph condensation methods provide a feasible path forward by synthesizing compact and representative datasets. Nevertheless, existing condensation approaches generally suffer from performance limitations in multimodal scenarios, mainly due to two reasons: (1) semantic misalignment between different modalities leads to gradient conflicts; (2) the message-passing mechanism of graph neural networks further structurally amplifies such gradient noise. Based on this, we propose Structural Regularized Gradient Matching (SR-GM), a condensation framework for multimodal graphs. This method alleviates gradient conflicts between modalities through a gradient decoupling mechanism and introduces a structural damping regularizer to suppress the propagation of gradient noise in the topology, thereby transforming the graph structure from a noise amplifier into a training stabilizer. Extensive experiments on four multimodal graph datasets demonstrate the effectiveness of SR-GM, highlighting its state-of-the-art performance and cross-architecture generalization capabilities in multimodal graph dataset condensation.
\end{abstract}

\ccsdesc[500]{Information systems~Data mining}
\ccsdesc[500]{Theory of computation~Graph algorithms analysis}
\keywords{Multimodal Graph Condensation,
Decoupled Gradient Matching,
Structural Regularization}
 
\maketitle

\section{Introduction}
The deep integration of the Internet and multimedia technologies has spawned a massive amount of graph-structured data containing rich heterogeneous information. From users in social networks and the multimedia content they publish \cite{khattar2019mvae,wang2012multimodal,su2021detecting,tao2020mgat,safavi2020codex} to product graphs associated with text descriptions and product images on e-commerce platforms \cite{xu2020product,deng2023construction}, these multimodal graphs have become the core of modern web applications. Graph neural networks (GNNs) \cite{kipf2016semi,rong2019dropedge,zhang2018link}, as powerful tools for processing such data, have achieved remarkable results in tasks such as node classification and link prediction. However, their success relies heavily on large-scale datasets \cite{safavi2020codex,wan2023scalable}, which leads to high computational costs, including lengthy training time, huge memory consumption, and complex hyperparameter tuning \cite{zhang2022efficient,cai2024multimodal}.

To address this challenge, a data-centric approach, dataset condensation \cite{wang2018dataset,zhao2020dataset}, has emerged. Its goal is to refine the original large-scale dataset into a small synthetic dataset with highly concentrated information, so that the model trained on this synthetic dataset can achieve performance comparable to that trained on the full dataset. In the field of computer vision, gradient matching \cite{zhao2020dataset} has become a benchmark method for achieving this goal. It efficiently learns highly representative synthetic samples by matching the model's training gradient trajectories on real data and synthetic data. Recently, researchers have begun to extend this paradigm to graph-structured data and proposed a variety of graph condensation methods \cite{jin2021graph,gao2023multiple,liu2022graph}. These methods are usually formulated as a complex bi-level optimization problem, aiming to learn a miniature synthetic graph structure and its node features through objectives such as gradient matching \cite{jin2021graph} or distribution matching \cite{liu2022graph}.

Although these methods show potential on unimodal graphs where node features come from a unimodal source, their performance degrades significantly when directly applied to more general multimodal graphs where nodes have multiple modal features, such as text and images. Through in-depth analysis, we find that the root cause lies in the conflict between multimodal gradient signals and the pathological amplification effect of the graph structure on this conflict. Specifically, there may be a semantic gap between different modalities \cite{liang2022mind,zhang2025beyond}, which leads to the generation of gradients in opposite directions during the optimization process. More importantly, the inherent message passing mechanism of GNN will quickly propagate and amplify the gradient conflicts of these local nodes to the entire graph, thus making the optimization process of the condensation algorithm based on gradient matching unstable and ineffective.

To overcome this problem, we propose a novel structurally regularized gradient matching (SR-GM) framework designed for multimodal graph condensation. The core ideas of SR-GM are twofold: (1) Gradient decoupling: We adopt the idea of conflict disentanglement from multi-task learning \cite{yu2020gradient,chai2024towards}. Specifically, for the conflicting gradients arising from different modalities on a synthetic node, we apply orthogonal projection to eliminate their inherent contradictions, thereby achieving synergistic inter-modal gradient collaboration; (2) Structural damping: On the propagation path, we innovatively introduce a graph Laplacian regularization term that acts on the gradient field rather than node features. This regularization term forces the optimization directions of adjacent nodes to be consistent during the condensation process, thereby transforming the graph structure from a "conflict amplifier" to an "optimization damper", effectively suppressing the spread of gradient noise.

We conducted extensive experiments on multiple multimodal graph benchmark datasets. Our results demonstrate that SR-GM significantly outperforms existing graph condensation baselines. Detailed ablation studies confirm the necessity and effectiveness of the two components, gradient decoupling and structural damping. Furthermore, the condensed graphs generated by SR-GM exhibit excellent generalization across diverse GNN architectures, highlighting their robustness and practicality.

The main contributions of this paper are as follows:
\begin{itemize}
\item We provide the first in-depth analysis of the mechanisms of gradient conflict in multimodal graph condensation and its negative effects amplified by graph structure, providing a novel theoretical perspective for understanding the underlying problem.
\item We propose the SR-GM method, which systematically addresses the core challenges of multimodal graph condensation through the dual design of gradient decoupling and structural damping.
\item We validate the superior performance of SR-GM through rigorous experiments and demonstrate the strong generalization of the generated condensed data in cross-architecture evaluations.
\end{itemize}

\begin{figure}
  \centering
  \begin{minipage}[b]{0.55\linewidth}
    \centering
    \includegraphics[width=\linewidth]{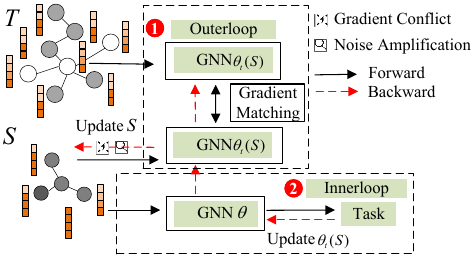}
    \\ \small (a)
  \end{minipage}
  \hfill
  \begin{minipage}[b]{0.44\linewidth}
    \centering
    \includegraphics[width=\linewidth]{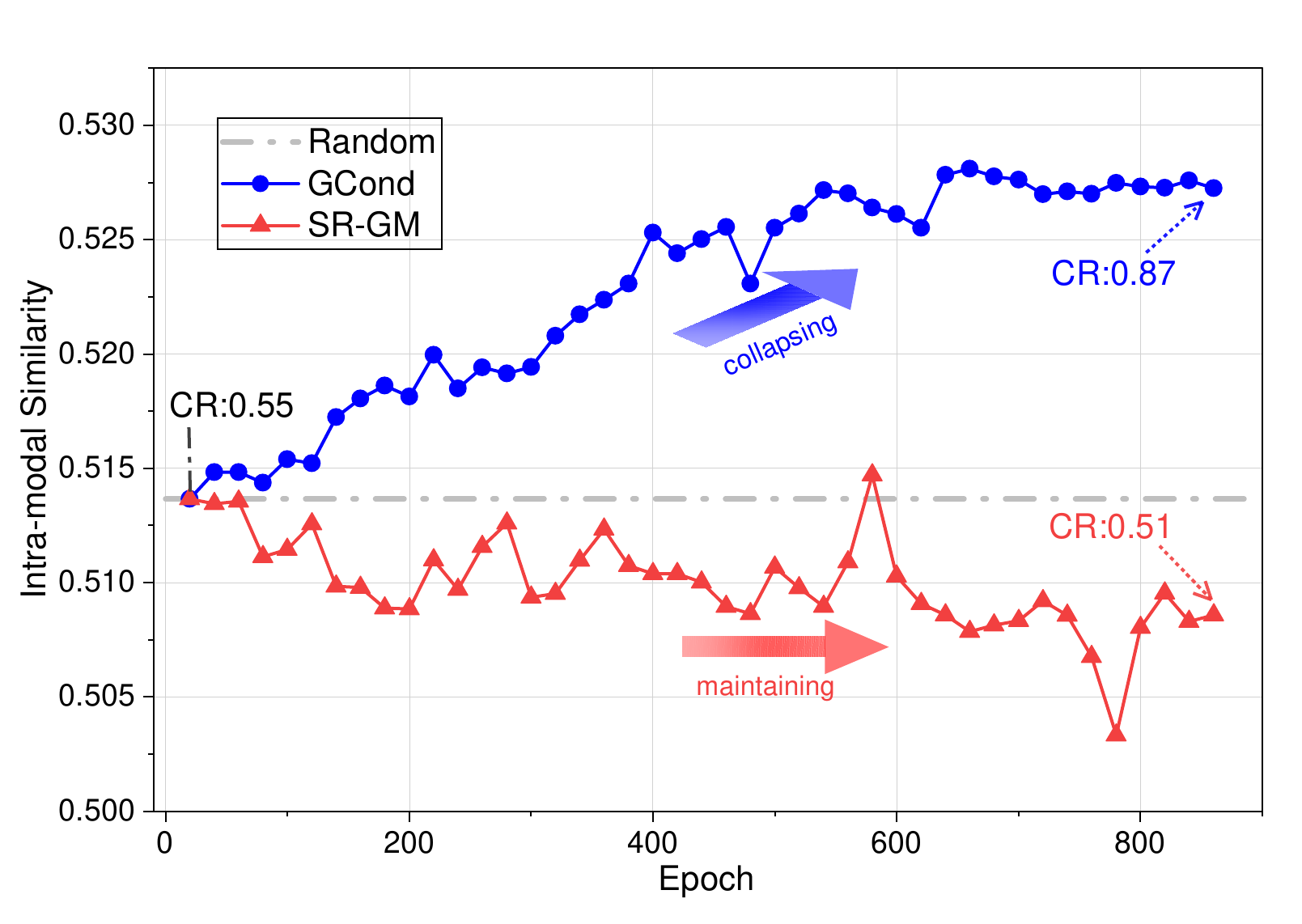}
    \\ \small  (b)
  \end{minipage}
  \caption[Impact of Multimodality]{
        (a): Gradient matching for multimodal Graph. (b): Intra-modal similarity trends during condensation. Gradient conflict and noise amplification cause modality collapse. In this process, text features move closer to visual features, which raises the intra-modal cosine similarity (blue curve). Features then cluster more tightly, driving the concentration ratio (CR)\footnotemark higher.
  }
  \label{fig:gm_multimodal}
\end{figure}
\footnotetext{A higher CR \cite{zhang2025beyond} indicates that the given similarity corresponds to a narrower directional region on the hypersphere, implying stronger feature concentration in the high-dimensional embedding space.}

\begin{figure*}[h]
\centering
\includegraphics[width=1.0\textwidth]{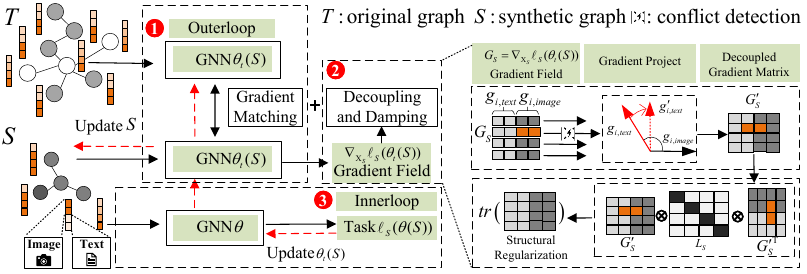}
\caption{This is the overall framework of SR-GM. $\mathcal{T}$ and $\mathcal{S}$ denote the target and condensed multimodal graph datasets, respectively. $\theta_t(\mathcal{S})$ denotes the parameters of the GNN trained on the synthetic graph $\mathcal{S}$, while $\nabla_{\mathbf{X}_\mathcal{S}} \ell_{\mathcal{S}} (\theta_t(\mathcal{S}) )$ denotes the gradient with respect to the synthetic node features $\mathbf{X}_\mathcal{S}$.}
\label{figure:framework}
\end{figure*}

\section{Methodology}

\subsection{Problem Formulation}

We consider the problem of multimodal graph condensation, where the objective is to distill a large-scale multimodal graph into a compact yet informative synthetic graph while preserving the essential information for downstream tasks.

Let the large-scale original graph be denoted as $\mathcal{T}=(\mathbf{X}, \mathbf{A}, \mathbf{Y})$, where $\mathbf{X} \in \mathbb{R}^{N \times d}$ is the node feature matrix for $N$ nodes, $\mathbf{A} \in \mathbb{R}^{N \times N}$ is the adjacency matrix, and $\mathbf{Y} \in \{0, 1, \dots, C-1\}^{N}$ are the node labels over $C$ classes. In the context of multimodal graphs, the $d$-dimensional feature vector for each node is a concatenation of features from different modalities, i.e., $\mathbf{X}_i = [\mathbf{x}_{i,\text{text}} \oplus \mathbf{x}_{i,\text{image}}]$.

The objective of graph condensation is to synthesize a small, information-rich graph $\mathcal{S} = (\mathbf{X}_{\mathcal{S}}, \mathbf{A}_{\mathcal{S}}, \mathbf{Y}_{\mathcal{S}})$ with $N'$ nodes ($N' \ll N$), such that a Graph Neural Network (GNN) model, $\phi_{\theta}$, trained on $\mathcal{S}$ achieves comparable performance to the same model trained on $\mathcal{T}$.

\subsection{Gradient Matching for Multimodal Graph }
To address the problem of multimodal graph dataset condensation, we adopt the graph dataset condensation paradigm. Specifically, the objective is formalized as a bi-level optimization problem \cite{jin2022condensing}. The outer loop aims to find the optimal synthetic graph $\mathcal{S}^{*}$ by minimizing a validation loss $\mathcal{L}_{\text{outer}}$ that evaluates a model trained on $\mathcal{S}$, while the inner loop trains the model on $\mathcal{S}$ to convergence to obtain the optimal parameters $\theta_{\mathcal{S}}^{*}$:

\begin{equation}
\mathcal{S}^{*} = \arg\min_{\mathcal{S}} \mathcal{L}_{\text{outer}}(\phi_{\theta_{\mathcal{S}}^{*}}, \mathcal{T}) \quad \text{s.t.} \quad \theta_{\mathcal{S}}^{*} = \arg\min_{\theta} \mathcal{L}_{\text{inner}}(\phi_{\theta}, \mathcal{S})
\end{equation}

Directly solving this bi-level optimization is computationally prohibitive, as each update to the synthetic graph $\mathcal{S}$ in the outer loop requires a complete inner-loop training of a GNN.

The Gradient Matching (GM) paradigm provides an effective approximation to this problem \cite{jin2021graph}. Rather than matching the performance of fully converged models, GM aligns the learning dynamics by matching the parameter gradients at each step along a single training trajectory. The process consists of two interleaved optimizations: an outer-loop update for the synthetic data $\mathcal{S}$ and an inner-loop update for the model parameters $\theta$. The objective is to minimize the cumulative distance between the gradients over the trajectory:

\begin{equation}
\min_{\mathcal{S}} \mathbb{E}_{\theta_0 \sim P_{\theta}} \left[ \sum_{t=0}^{T-1} D(\nabla_{\theta} \mathcal{L}_{\mathcal{S}}(\theta_t), \nabla_{\theta} \mathcal{L}_{\mathcal{T}}(\theta_t)) \right]
\end{equation}

Here, model parameters are updated in the inner loop using the synthetic data: $\theta_{t+1} = \theta_t - \eta_{\theta} \nabla_{\theta} \mathcal{L}_{\mathcal{S}}(\theta_t)$, where $\mathcal{L}_{\mathcal{S}}$ and $\mathcal{L}_{\mathcal{T}}$ denote the task losses on the synthetic and original graphs, respectively, and $D(\cdot,\cdot)$ is a distance metric between gradient vectors.

Although traditional gradient matching methods succeed in unimodal settings, they struggle on multimodal graphs due to gradient conflicts and structural noise amplification. Our analysis Fig.~\ref{fig:gm_multimodal} shows that standard gradient matching compresses inter-modal differences during training, leading to modality collapse—where multimodal information degenerates into a unimodal representation. Moreover, GNN message passing amplifies gradient noise, correlating update directions across nodes and accelerating feature concentration (higher CR). In the following sections, we will analyze these two issues in detail.

\subsection{Inter-Modality Gradient Conflict}

In the gradient matching process for multimodal graph condensation, directional conflicts occur between the feature gradients of different modalities within a node, the root of which can be traced back to the inherent opposition of modal gradients in the parameter space. The parameter gradient of the original graph, \(\nabla_\theta \mathcal{L}_T\), can be decomposed into independent components for the text and image modalities, \(v_t\) and \(v_i\). When the optimal update directions of the two modalities are inconsistent, the angle between them satisfies \(\psi > \pi/2\), resulting in an inner product \(\langle v_t, v_i \rangle < 0\), thereby forming a modal conflict at the parameter level.

Gradient matching aims to make the synthetic gradient \(\nabla_\theta \mathcal{L}_S\) approximate this target, i.e., \(\nabla_\theta \mathcal{L}_S \approx \alpha \nabla_\theta \mathcal{L}_T\). The synthetic gradient is connected to the node feature gradients via the Jacobian matrices \(A_i\) and \(B_i\).

Specifically, let the text feature of node \(i\) be \(x_{i,\text{text}}\) and its image feature be \(x_{i,\text{image}}\). Their corresponding feature gradients are \(g_{i,\text{text}} = \frac{\partial \mathcal{L}_S}{\partial x_{i,\text{text}}}\) and \(g_{i,\text{image}} = \frac{\partial \mathcal{L}_S}{\partial x_{i,\text{image}}}\), respectively. The Jacobian matrices \(A_i = \frac{\partial}{\partial x_{i,\text{text}}} \nabla_\theta \mathcal{L}_S \in \mathbb{R}^{|\theta| \times d_t}\) and \(B_i = \frac{\partial}{\partial x_{i,\text{image}}} \nabla_\theta \mathcal{L}_S \in \mathbb{R}^{|\theta| \times d_v}\) represent the partial derivatives of the parameter gradient with respect to the node's text and image features, respectively. They linearly map the node feature gradients into the parameter gradient space. Therefore, the synthetic gradient can be expressed as:

\begin{equation}
\label{eq:synthetic_gradient}
\nabla_\theta \mathcal{L}_S = \sum_{i=1}^{N'} \left( A_i \, g_{i,\text{text}} + B_i \, g_{i,\text{image}} \right)
\end{equation}

Due to parameter sharing and feature fusion, the column spaces of \(A_i\) and \(B_i\) often overlap (modal mixing). To satisfy the matching condition \(A_i \, g_{i,\text{text}} + B_i \, g_{i,\text{image}} \approx \alpha (v_t + v_i)\), the optimization process must reconcile the opposing directions of \(v_t\) and \(v_i\) within the overlapping subspace. This forces the projections of the feature gradients \(g_{i,\text{text}}\) and \(g_{i,\text{image}}\) onto this subspace to become opposite in direction, thereby triggering intra-node feature gradient conflict:

\begin{equation}
\label{eq:gradient_conflict}
\langle g_{i,\text{text}}, g_{i,\text{image}} \rangle < 0
\end{equation}

Therefore, intra-node modal feature gradient conflict is an inevitable outcome when gradient matching seeks to fit a multimodal target that is inherently conflicted, under a model structure with modal mixing. The key to addressing this issue lies in gradient decoupling. As shown in Fig.~\ref{fig:gm_multimodal}(b), decoupling such conflict effectively preserves inter-modal feature diversity and prevents collapse.

\subsection{Structural Amplification of Gradient Noise}

The meta-gradient calculation in gradient matching involves backpropagation through the GNN architecture, which amplifies the noise in modal gradients. In the outer loop, the update direction of the synthetic graph is given by:
\begin{equation}\label{eq:meta_gradient_update}
\Delta \mathbf{X}_\mathcal{S} = -\eta \nabla_{\mathbf{X}_S} D(\nabla_\theta \mathcal{L}_\mathcal{S}, \nabla_\theta \mathcal{L}_T)
\end{equation}

This meta-gradient is computed via the chain rule, which requires backpropagation through the message-passing layers of the GNN. Specifically, during meta-gradient computation, the gradient of a node's features is influenced by the gradients of its neighbors, causing gradients to propagate backward along the graph edges:
\begin{equation}\label{eq:gradient_backpropagation}
\nabla_{h_i^{(k-1)}} \mathcal{L} = \sum_{j \mid i \in \mathcal{N}(j)} \frac{\partial \mathcal{L}}{\partial h_j^{(k)}} \frac{\partial h_j^{(k)}}{\partial h_i^{(k-1)}}.
\end{equation}

This dynamic transforms the computational graph of the GNN into a medium for noise propagation. The gradient conflicts at local nodes are not contained; instead, they are amplified and propagated throughout the graph structure. Let the gradient noise of node representations at the \(l\)-th layer be denoted by \(\mathbf{R}^{(l)}\). Then the noise propagation in backpropagation can be formalized as:
\begin{equation}\label{eq:noise_propagation}
\mathbf{R}^{(l-1)} = \mathbf{A}_S^\top \mathbf{R}^{(l)} \mathbf{W}^{(l)\top} + \mathbf{E}^{(l)},
\end{equation}
where \(\mathbf{A}_S\) is the adjacency matrix of the synthetic graph, \(\mathbf{W}^{(l)}\) is the weight matrix of the \(l\)-th layer, and \(\mathbf{E}^{(l)}\) is the inherent modal mixing noise at that layer. This results in a highly non-smooth gradient field over the synthetic graph, where the optimization directions of adjacent nodes can differ significantly.

To quantify this effect, we define a noise amplification operator on the graph, \(\mathcal{P}: \mathbb{R}^{N' \times d} \to \mathbb{R}^{N' \times d}\), as \(\mathcal{P}(\mathbf{R}) = \mathbf{L}_S \mathbf{R}\), where \(\mathbf{L}_S\) is the Laplacian matrix of the synthetic graph and \(\mathbf{R}\) is the node gradient noise matrix. This operator captures the amplification effect of the graph structure on noise. Based on this, we obtain the following theorem:

\begin{theorem}\label{thm:noise_amplification}
Under gradient matching optimization, the amplification factor of modal mixing noise propagated through the graph structure is bounded by the Dirichlet energy of the gradient field:
\begin{equation}\label{eq:noise_amplification_bound}
\|\mathbf{L}_S \mathbf{R}\|_F^2 \leq \lambda_{\max}(\mathbf{L}_S) \cdot E(\mathbf{R}),
\end{equation}
where \(E(\mathbf{R}) = \operatorname{tr}(\mathbf{R}^\top \mathbf{L}_S \mathbf{R})\) is the Dirichlet energy of the gradient field, and \(\lambda_{\max}(\mathbf{L}_S)\) is the maximum eigenvalue of the Laplacian matrix.
\end{theorem}

The proof of Theorem~\ref{thm:noise_amplification} is provided in Appendix \ref{proof_t}. Theorem~\ref{thm:noise_amplification} shows that the graph structure systematically amplifies gradient noise through its spectral properties, thereby exacerbating optimization instability. In particular, a high Dirichlet energy corresponds to a chaotic gradient field, which the structural amplification mechanism sustains, resulting in a rugged optimization landscape and destabilized meta-gradient. This persistent instability directly hinders the condensation objective.

\begin{figure}
  \centering
  \begin{minipage}[b]{0.49\linewidth}
    \centering
    \includegraphics[width=\linewidth]{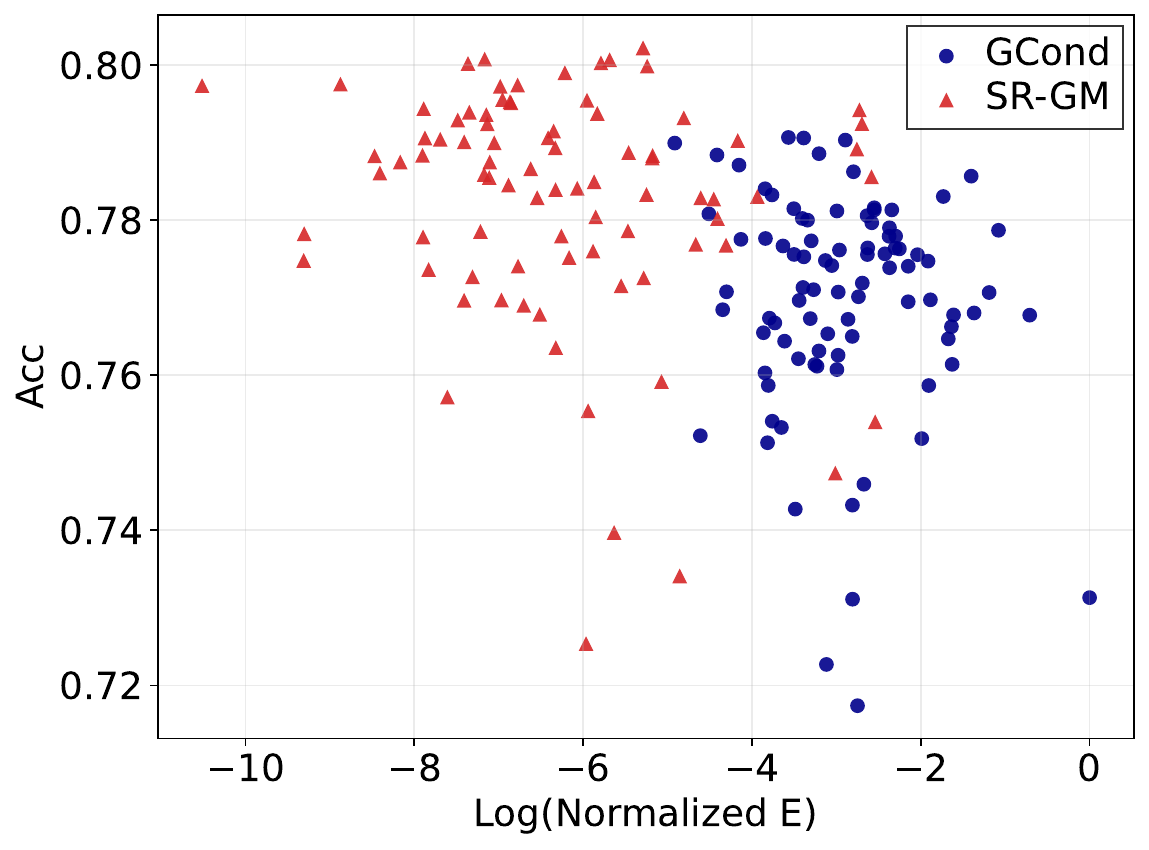}
    \\ \small (a)
  \end{minipage}
  \hfill
  \begin{minipage}[b]{0.49\linewidth}
    \centering
    \includegraphics[width=\linewidth]{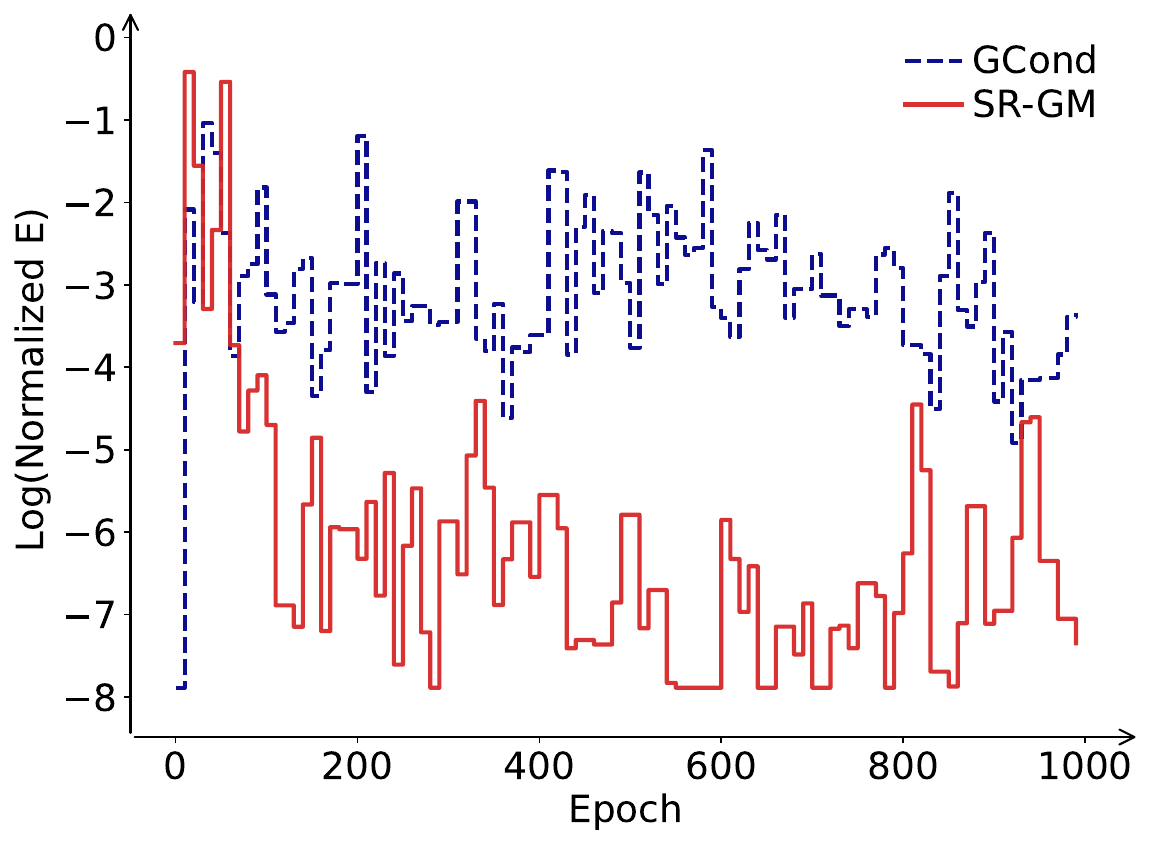}
    \\ \small (b)
  \end{minipage}
  \caption[structure noise]{
        Performance and Dirichlet energy\footnotemark analysis. (a): Relationship between node classification accuracy and $E(\mathbf{R})$ during a condensation process; (b): Evolutionary trends of $E(\mathbf{R})$ for GCond and SR-GM. For clarity, $E(\mathbf{R})$ is normalized and log-transformed.
  }
  \label{fig:strut_noise_analysis}
\end{figure}

We further analyze the impact of Dirichlet energy \(E(\mathbf{R})\) on gradient field stability and condensation performance through experiments. The results confirm a clear negative correlation Fig.~\ref{fig:strut_noise_analysis}(a): lower \(E(\mathbf{R})\) consistently leads to better condensed graph performance. Thus, suppressing this energy field stabilizes the gradient field and enhances condensation performance, as shown in Fig.~\ref{fig:strut_noise_analysis}(b).

\footnotetext{Dirichlet energy \cite{zhou2021dirichlet} is a key metric that measures the smoothness of graph signals or embeddings over the graph topology.}

\subsection{The Proposed SR-GM Framework}

To address this dual challenge, as shown in Fig.~\ref{figure:framework}, we propose \textbf{Structurally-Regularized Gradient Matching (SR-GM)}, a framework that re-engineers the condensation process through two synergistic components: Gradient Decoupling and Structural Damping.
\subsubsection{Parametrization of the Synthetic Graph}
In SR-GM, the synthetic graph $\mathcal{S}$ is parameterized by its node features $\mathbf{X}_{\mathcal{S}}$ and the parameters $\Phi$ of a structure generator network $g_{\Phi}$. The node features $\mathbf{X}_{\mathcal{S}}$ are learned directly. Following GCond~\cite{jin2021graph}, the adjacency matrix $\mathbf{A}_{\mathcal{S}}$ is not directly parameterized but is generated as a function of the node features:
$
\mathbf{A}_{\mathcal{S}} = g_{\Phi}(\mathbf{X}_{\mathcal{S}})
$.

Specifically, the weight of an edge between nodes $i$ and $j$ is computed by a shared MLP, which is a multi-layer perceptron with parameters $\Phi$:

\begin{equation}
(\mathbf{A}_{\mathcal{S}})_{ij} = \text{Sigmoid}\left( \frac{\text{MLP}_{\Phi}([\mathbf{x}_i; \mathbf{x}_j]) + \text{MLP}_{\Phi}([\mathbf{x}_j; \mathbf{x}_i])}{2} \right)
\label{eq:structure_generation}
\end{equation}

where $[\cdot;\cdot]$ denotes concatenation. This formulation makes the graph structure differentiable with respect to both the node features $\mathbf{X}_{\mathcal{S}}$ and the generator parameters $\Phi$, and naturally enforces symmetry for undirected graphs. The set of learnable parameters for the outer loop is thus $\{\mathbf{X}_{\mathcal{S}}, \Phi\}$.

\subsubsection{Gradient Decoupling via Orthogonal Projection}
To resolve the root cause of instability, we perform "gradient surgery" on the feature gradients \cite{liu2021conflict,yu2020gradient} of each synthetic node. This process is a prerequisite for computing the structural regularization term and is integrated into the outer loop's meta-gradient calculation.

\textbf{Working Mechanism:} At each inner-loop step $t$, after computing the synthetic loss $\mathcal{L}_{\mathcal{S}}(\theta_t)$, we first compute the gradient of this loss with respect to the synthetic features $\mathbf{X}_{\mathcal{S}}$. This yields the feature gradient matrix $\mathbf{G}_{\mathcal{S}} \in \mathbb{R}^{N' \times d}$:

\begin{equation}
\mathbf{G}_{\mathcal{S}} = \nabla_{\mathbf{X}_{\mathcal{S}}} \mathcal{L}_{\mathcal{S}}(\theta_t)
\label{eq:feat_gradient}
\end{equation}

Since the synthetic features $\mathbf{X}_{\mathcal{S}}$ are a concatenation of modal features, the resulting gradient matrix $\mathbf{G}_{\mathcal{S}}$ has the same structure. For each node $s_i$, we can partition its corresponding row vector in $\mathbf{G}_{\mathcal{S}}$ to obtain the modality-specific gradients, $\mathbf{g}_{i,\text{text}}$ and $\mathbf{g}_{i,\text{image}}$.

If a conflict exists ($\langle \mathbf{g}_{i,\text{text}}, \mathbf{g}_{i,\text{image}} \rangle < 0$), we project each gradient onto the normal plane of the other to eliminate the conflicting components, following the PCGrad \cite{yu2020gradient} methodology:

\begin{align}
\mathbf{g'}_{i,\text{text}} &= \mathbf{g}_{i,\text{text}} - \frac{\langle \mathbf{g}_{i,\text{text}}, \mathbf{g}_{i,\text{image}} \rangle}{\|\mathbf{g}_{i,\text{image}}\|^2} \mathbf{g}_{i,\text{image}} \label{eq:decouple_text} \\
\mathbf{g'}_{i,\text{image}} &= \mathbf{g}_{i,\text{image}} - \frac{\langle \mathbf{g}_{i,\text{image}}, \mathbf{g}_{i,\text{text}} \rangle}{\|\mathbf{g}_{i,\text{text}}\|^2} \mathbf{g}_{i,\text{text}} \label{eq:decouple_image}
\end{align}

This operation yields a decoupled feature gradient matrix $\mathbf{G'}_{\mathcal{S}}$, where each row is the reconciled gradient $\mathbf{g'}_i = [\mathbf{g'}_{i,\text{text}} \oplus \mathbf{g'}_{i,\text{image}}]$.

\subsubsection{Structural Damping via Gradient Field Regularization}

To counteract the structural amplification of noise, we introduce a novel regularization term that operates directly on the decoupled gradient field. Unlike traditional graph regularization \cite{yang2021rethinking} on node features, which GNNs already implicitly perform, our regularizer enforces smoothness on the optimization directions themselves. The structural regularization term is the Dirichlet energy of this gradient field:

\begin{equation}
\mathcal{R}_{\text{struct}}(\mathbf{G'}_{\mathcal{S}}) = \mathrm{tr}(\mathbf{G'}_{\mathcal{S}}^{\top} \mathbf{L}_{\mathcal{S}} \mathbf{G'}_{\mathcal{S}}) = \sum_{(i,j) \in E_{\mathcal{S}}} w_{ij} \|\mathbf{g'}_i - \mathbf{g'}_j\|^2
\label{eq:structure_match_loss}
\end{equation}

This term penalizes large differences in optimization directions between connected nodes, forcing the gradient field to be smooth. It thereby transforms the graph structure from a "conflict amplifier" into an "optimization damper."

\subsubsection{Final Objective and Algorithm}

By integrating both components, the final SR-GM objective function becomes:

\begin{equation}
\min_{\mathcal{S}} \mathbb{E}_{\theta_t} \left[ D(\nabla_{\theta} \mathcal{L}_{\mathcal{S}}(\theta_t), \nabla_{\theta} \mathcal{L}_{\mathcal{T}}(\theta_t)) + \lambda \cdot \mathcal{R}_{\text{struct}} \right] 
\end{equation}

where $\lambda$ is a hyperparameter balancing gradient matching accuracy with gradient field smoothness.The gradient matching loss $\mathcal{L}_{\text{gm}} = D\left(\nabla_{\theta} \mathcal{L}_{\mathcal{S}}(\theta_t), \nabla_{\theta} \mathcal{L}_{\mathcal{T}}(\theta_t)\right)$ follows the implementation in GCond \cite{jin2021graph}. SR-GM design fundamentally repurposes the graph structure from a source of instability into a tool for stabilization. 
The detailed algorithm of SR-GM is shown in Algorithm~\ref{alg:sr-gm}.

\begin{algorithm}
\caption{Structure-Regularized Gradient Matching (SR-GM)}
\label{alg:sr-gm}
\begin{algorithmic}[1]
\Require Original graph $\mathcal{T}$, GNN model $\phi_{\theta}$, hyperparameters $\lambda$, $\eta_{\mathcal{S}}$, $\eta_{\Phi}$, $\eta_{\theta}$, outer-loop iterations $K$, inner-loop iterations $T$
\Ensure Optimized synthetic features $\mathbf{X}_{\mathcal{S}}$ and structure generator $g_{\Phi}$
\State Initialize synthetic features $\mathbf{X}_{\mathcal{S}}$ and $g_{\Phi}$ parameters $\Phi$
\For{$k = 1$ \textbf{to} $K$}
    \State Initialize GNN parameters $\theta_0 \sim P_{\theta_0}$
    \For{$t = 0$ \textbf{to} $T-1$}
        \State Sample a batch $\mathcal{T}'$ from $\mathcal{T}$
        \State Compute real data gradient $\mathbf{g}_{\mathcal{T}} = \nabla_{\theta} \mathcal{L}_{\mathcal{T}'}(\theta_t)$
        \State Generate synthetic adjacency matrix $\mathbf{A}_{\mathcal{S}} = g_{\Phi}(\mathbf{X}_{\mathcal{S}})$ with Eq.~(\ref{eq:structure_generation})
        \State Compute synthetic loss $\mathcal{L}_{\mathcal{S}}(\theta_t)$ on $\mathcal{S} = (\mathbf{X}_{\mathcal{S}}, \mathbf{A}_{\mathcal{S}})$
        \State Compute synthetic model gradient $\mathbf{g}_{\mathcal{S}} = \nabla_{\theta} \mathcal{L}_{\mathcal{S}}(\theta_t)$
        \State Compute synthetic feature gradient matrix $\mathbf{G}_{\mathcal{S}} = \nabla_{\mathbf{X}_{\mathcal{S}}} \mathcal{L}_{\mathcal{S}}(\theta_t)$
        
        \Statex \textbf{// Component I: Gradient Decoupling}
        \State Construct decoupled gradient matrix $\mathbf{G}'_{\mathcal{S}}$ from $\mathbf{G}_{\mathcal{S}}$ with Eq.~(\ref{eq:decouple_text})(\ref{eq:decouple_image})
        
        \Statex \textbf{// Component II: Structural Damping}
        \State Compute structural regularization loss $\mathcal{R}_{\text{struct}}$ with Eq.~(\ref{eq:structure_match_loss})
        
        \Statex \textbf{// Meta-Update of Synthetic Data (Outer Loop)}
        \State Compute gradient matching loss $\mathcal{L}_{\text{gm}} = D(\mathbf{g}_{\mathcal{S}}, \mathbf{g}_{\mathcal{T}})$
        \State Compute total update loss $\mathcal{L}_{\text{update}} = \mathcal{L}_{\text{gm}} + \lambda \cdot \mathcal{R}_{\text{struct}}$
        \State Update $\mathbf{X}_{\mathcal{S}}$, $\Phi$ using $\nabla_{\mathbf{X}_{\mathcal{S}}, \Phi} \mathcal{L}_{\text{update}}$ and learning rates $\eta_{\mathcal{S}}$, $\eta_{\Phi}$
        
        \Statex \textbf{// Update Model Parameters (Inner Loop Curriculum Step)}
        \State Update model parameters $\theta_{t+1} \leftarrow \theta_t - \eta_{\theta} \mathbf{g}_{\mathcal{S}}$
    \EndFor
\EndFor
\State \Return Optimized synthetic features $\mathbf{X}_{\mathcal{S}}$ and generator parameters $\Phi$
\end{algorithmic}
\end{algorithm}

\section{Experiment}
To verify the effectiveness and robustness of the SR-GM framework, we design comprehensive experiments to explore the following research questions: \textbf{Q1}: Can SR-GM outperform the other graph condensation methods? \textbf{Q2}: How does SR-GM perform under scenarios where multimodal features are unaligned? \textbf{Q3}: Can SR-GM offer competitive condensation efficiency? \textbf{Q4}: How do gradient decoupling and structural regularization individually and synergistically impact the condensation quality? \textbf{Q5}: Can SR-RM leverage the complementary information from multimodal features? \textbf{Q6}: How do different downstream settings and hyperparameters affect SR-GM's generalizability? Due to space constraints, additional experimental results are provided in the Appendix~\ref{appendix}. All the experiments are conducted on an NVIDIA RTX 3090 GPU.

\begin{table}
    \centering
    \caption{Statistics of datasets.}
    \label{tab:data_stat}
    \begin{tabular}{lccc}
    \toprule
    Name & Nodes & Edges & Train/Val/Test \\
    \midrule
    Ele-fashion & 97,766 & 199,602 & 58,659/9,777/29,330 \\
    Goodreads-NC & 685,294 & 7,235,084 & 406,689/67,805/203,428 \\
    Amazon-Sports & 31,073 & 351,294 & 21,751/4,660/4,662 \\
    Amazon-Cloth & 84,357 & 1,107,964 & 59,049/12,653/12,655 \\
    \bottomrule
\end{tabular}
\end{table}

\begin{table*}
    \caption{Inductive performance on Ele-fashion , Goodreads-NC  and Transductive performance on Amazon-Sports, Amazon-Cloth. Performance is reported as test accuracy (\%) for inductive and AUC score for transductive. "Original" denotes models trained on original graph. The best accuracies are bolded and the second best are underlined. $\triangle$(\%) denotes the improvements of SR-GM upon the sub-optimal result.}
    \label{tab:baseline_comp}
    \centering
    \small
    \setlength{\tabcolsep}{2pt}
    \begin{tabular}{lccccccccccc} 
    \toprule
    \multicolumn{2}{c}{} & \multicolumn{7}{c}{Baselines} & \multicolumn{2}{c}{Proposed} & \multicolumn{1}{c}{} \\
    \cmidrule(lr){3-9} \cmidrule(lr){10-11}
    \raisebox{1.5ex}[0pt]{Dataset} & \raisebox{1.5ex}[0pt]{Ratio ($r$)} & Random & Herding & K-Center & MSGC & SGDD & GCDM & GCond & SR-GM & $\triangle$(\%) & \raisebox{1.5ex}[0pt]{Original} \\
    \midrule
    & 0.100\% & 63.28$\pm$0.25 & 64.17$\pm$0.16 & 65.59$\pm$0.21 & 68.71$\pm$0.82 & 71.08$\pm$0.08 & 76.60$\pm$0.09 & \underline{78.93$\pm$0.06} & \textbf{79.85$\pm$0.22} & $\uparrow$0.92 & \\
    Ele-fashion & 0.300\% & 68.48$\pm$0.49 & 67.40$\pm$0.50 & 68.82$\pm$0.64 & \underline{76.93$\pm$0.37} & 75.68$\pm$0.62 & 76.21$\pm$0.06 & 68.00$\pm$0.78 & \textbf{82.03$\pm$0.13} & $\uparrow$5.10 & 85.68$\pm$0.08 \\
    & 0.500\% & 70.25$\pm$0.24 & 67.41$\pm$0.14 & 70.37$\pm$0.36 & \underline{79.09$\pm$0.40} & 75.15$\pm$0.11 & 77.19$\pm$0.12 & 71.42$\pm$0.28 & \textbf{80.64$\pm$0.17} & $\uparrow$1.55 & \\
    \midrule 
    & 0.025\% & 46.45$\pm$0.47 & 44.71$\pm$0.32 & 44.78$\pm$0.77 & 38.14$\pm$1.43 & 36.75$\pm$0.25 & 42.08$\pm$0.32 & \underline{57.59$\pm$0.55} & \textbf{66.43$\pm$0.60} & $\uparrow$8.84 & \\
    \raisebox{1.5ex}[0pt]{Goodreads-NC} & 0.050\% & 47.60$\pm$0.27 & 44.73$\pm$0.36 & 44.82$\pm$0.35 & 43.10$\pm$0.23 & 31.68$\pm$1.27 & 43.56$\pm$0.43 & \underline{58.09$\pm$0.66} & \textbf{68.20$\pm$0.31} & $\uparrow$10.11 & \raisebox{1.5ex}[0pt]{79.45$\pm$0.04} \\
    \midrule
    & 0.100\% & 61.14$\pm$0.14 & 61.13$\pm$0.15 & 62.10$\pm$0.20 &  \underline{65.43$\pm$0.53} & 63.52$\pm$0.70 & 64.30$\pm$0.19 & 64.14$\pm$0.53 & \textbf{67.78$\pm$0.52} & $\uparrow$2.35 & \\
    Amazon-Sports & 0.300\% & 59.23$\pm$0.63 & 61.51$\pm$0.07 & 61.76$\pm$0.58 & \underline{65.07$\pm$0.53} & 63.07$\pm$0.42 & 64.55$\pm$0.57 & 64.86$\pm$0.75 & \textbf{70.25$\pm$0.55} & $\uparrow$5.18 & 70.50$\pm$0.47 \\
    & 0.500\% & 60.37$\pm$0.25 & 61.76$\pm$0.25 & 64.31$\pm$0.24 & \underline{67.90$\pm$0.12} & 65.56$\pm$0.42 & 64.23$\pm$0.48 & 65.28$\pm$0.47 & \textbf{71.12$\pm$0.70} & $\uparrow$3.22 &  \\
    \midrule
    & 0.100\% & 52.49$\pm$0.01 & 50.82$\pm$0.15 & 51.11$\pm$0.20 & 57.80$\pm$0.30 & \underline{58.30$\pm$0.49} & 56.99$\pm$0.40 & 57.46$\pm$0.57 & \textbf{65.96$\pm$0.44} & $\uparrow$7.66 &  \\
    Amazon-Cloth & 0.300\% & 54.85$\pm$0.46 & 53.31$\pm$0.35 & 57.26$\pm$0.48 & \underline{65.88$\pm$0.22} & 61.47$\pm$0.07 & 56.97$\pm$0.43 & 58.32$\pm$0.67 & \textbf{66.87$\pm$0.30} & $\uparrow$0.99 & 71.59$\pm$0.27 \\
    & 0.500\% & 52.92$\pm$0.23 & 54.54$\pm$0.50 & 53.75$\pm$0.36 & \underline{64.97$\pm$0.14} & 61.37$\pm$0.48 & 56.74$\pm$0.41 & 57.88$\pm$0.99 & \textbf{67.64$\pm$0.14} & $\uparrow$2.67 &  \\
    \bottomrule
    \end{tabular}
\end{table*}

\subsection{Experimental Setup}

\paragraph{Datasets}
We evaluate the condensation performance of the proposed framework on 4 multimodal datasets, i.e., Ele-fashion \cite{ni2019justifying} and Goodreads-NC \cite{wan2018item} for inductive setting and Amazon-Sports, Amazon-Cloth for transductive setting. Following \cite{cai2024multimodal}, we construct binary classification datasets for Amazon-Sports and Amazon-Cloth  from the Amazon e-commerce platform\footnote{\url{https://www.amazon.com/}}; reviews with ratings $\geq$ 4 are labeled as positive, and those with ratings < 2 as negative. Text and visual features for the Amazon-Sports and Amazon-Cloth datasets are encoded and aligned with ImageBind \cite{girdhar2023imagebind}, while CLIP \cite{radford2021learning} is used for the Goodreads-NC and Ele-fashion datasets. The resulting features have dimensions of 1024 for Ele-fashion and Goodreads-NC, and 2048 for Amazon-Sports and Amazon-Cloth, and in all cases, the text and image modalities each constitute half of the total dimensionality.  For the inductive setting, the test graph is not available during training. Dataset statistics are shown in Table~\ref{tab:data_stat}.

\paragraph{Baselines}
We benchmark SR-GM against a comprehensive suite of baselines. These methods can be categorized into \textbf{coreset selection} and \textbf{graph condensation} methods. For \textbf{coreset selection}, we induce a subgraph from a representative node subset. We evaluate three selection strategies: \textbf{Random}, \textbf{Herding} \cite{welling2009herding}, and \textbf{K-Center} \cite{sener2017active}. For \textbf{graph condensation}, we employ four methods. \textbf{GCDM} \cite{liu2022graph} directly aligns the class distributions of  the original and condensed graphs. \textbf{GCond} \cite{jin2021graph} and \textbf{MSGC} \cite{gao2023multiple} aim to preserve performance by matching GNN training gradients. In contrast, \textbf{SGDD} \cite{yang2023does} focus on maintaining global graph properties by preserving the graph's spectral characteristics and connectivity patterns.

\paragraph{Evaluation}
We evaluate the quality of the condensed graphs by their ability to preserve performance on the downstream task of node classification. For each baseline and our proposed SR-GM, we first generate a condensed graph containing $rN$ nodes from the original training graph of $N$ nodes, where $r \in (0, 1)$ is the condensation ratio.
The evaluation follows a two-stage protocol. In the \textbf{training stage}, a standard GNN model is trained exclusively on the small, condensed graph. Crucially, the original full graph is entirely discarded and inaccessible during this stage. In the subsequent \textbf{test stage}, the GNN model trained on the condensed graph is applied to the original, unseen test graph to perform inference on the test nodes. The primary metric for evaluation is the resulting test accuracy. To ensure robust conclusions, we repeat all experiments 5 times and report the mean accuracy and its variance.

\subsection{Comparison with Baselines (Q1)}
In this subsection, we test the performance of a 2-layer GCN on the condensed graphs, and compare the proposed SR-GM with baselines. Table~\ref{tab:baseline_comp} reports node classification performance. As shown in Table~\ref{tab:baseline_comp}, we make the following observations:

\begin{description}
\item[Superiority of SR-GM.]
SR-GM outperforms most baseline methods across all datasets and condensation ratios. Notably, on the Goodreads-NC dataset (0.025\% ratio), SR-GM achieves 66.43\% accuracy, surpassing the strongest baseline, GCond, by nearly 9 percentage points.
\item[Stability and Robustness.]
Beyond accuracy, SR-GM demonstrates exceptional reliability. Unlike SGDD and MSGC, which fail to surpass basic coreset methods on Goodreads-NC (Table~\ref{tab:baseline_comp}), SR-GM consistently maintains high performance with low variance.
\item[Impact of Condensation Ratio.]
While increasing the condensation ratio typically boosts performance by expanding information capacity, it also makes optimization harder. Despite these challenges, SR-GM consistently outperforms other methods. 
\end{description}

\begin{table*}
    \caption{Inductive performance on Ele-fashion, Goodreads-NC and Transductive performance on Amazon-Sports, Amazon-Cloth using T5-ViT encoding. Performance is reported as test accuracy (\%) for inductive and AUC score for transductive. "Original" denotes models trained on original graph. The best accuracies are bolded and the second best are underlined. $\triangle$(\%) denotes the improvements of SR-GM upon the sub-optimal result.}
    \label{tab:baseline_comp_t5vit}
    \centering
    \small
    \setlength{\tabcolsep}{2pt}
    \begin{tabular}{lccccccccccc} 
    \toprule
    \multicolumn{2}{c}{} & \multicolumn{7}{c}{Baselines} & \multicolumn{2}{c}{Proposed} & \multicolumn{1}{c}{} \\
    \cmidrule(lr){3-9} \cmidrule(lr){10-11}
    \raisebox{1.5ex}[0pt]{Dataset} & \raisebox{1.5ex}[0pt]{Ratio ($r$)} & Random & Herding & K-Center & MSGC & SGDD & GCDM & GCond & SR-GM & $\triangle$(\%) &\raisebox{1.5ex}[0pt]{Original} \\
    \midrule
    & 0.100\% & 68.09$\pm$0.81 & 67.23$\pm$0.30 & 65.25$\pm$0.35 & \underline{73.03$\pm$0.11}& 72.85$\pm$0.14 & 71.98$\pm$0.27 & 71.44$\pm$0.22& \textbf{74.09$\pm$0.13} & $\uparrow$1.06& \\
    Ele-fashion & 0.300\% & 72.22$\pm$0.18 & 70.77$\pm$0.34 & 70.26$\pm$0.21 & \underline{77.28$\pm$0.08}& 72.75$\pm$0.33 & 73.88$\pm$0.11 & 75.21$\pm$0.11& \textbf{77.47$\pm$0.14} & $\uparrow$0.19 & 84.47$\pm$0.05 \\
    & 0.500\% & 70.83$\pm$0.79 & 69.07$\pm$0.54 & 71.93$\pm$0.50 & \underline{76.19$\pm$0.22}& 73.18$\pm$0.45 & 73.83$\pm$0.08 & 75.59$\pm$0.25& \textbf{78.23$\pm$0.13} & $\uparrow$2.04 & \\
    \midrule 
    & 0.025\% & 38.86$\pm$1.07 & 41.43$\pm$0.70 & 36.34$\pm$0.78 & \underline{50.37$\pm$0.42} & 48.95$\pm$1.08 & 44.08$\pm$0.83 & 40.53$\pm$0.30& \textbf{55.28$\pm$0.11} & $\uparrow$4.91& \\
    \raisebox{1.5ex}[0pt]{Goodreads-NC} & 0.050\% & 44.48$\pm$0.01 & 46.91$\pm$0.92 & 40.35$\pm$0.54 & 52.95$\pm$0.15& 52.87$\pm$0.29 & 46.11$\pm$0.95 & \underline{53.22$\pm$0.18}& \textbf{61.03$\pm$0.21} & $\uparrow$7.81& \raisebox{1.5ex}[0pt]{76.93$\pm$0.02} \\
    \midrule
    & 0.100\% & 60.29$\pm$0.08 & 61.22$\pm$0.16 & 58.86$\pm$0.16 & \underline{68.06$\pm$0.08}& 64.79$\pm$0.11 & 66.24$\pm$0.43 & 67.87$\pm$0.19& \textbf{71.95$\pm$0.18} & $\uparrow$3.89 & \\
    Amazon-Sports & 0.300\% & 59.00$\pm$0.17 & 60.56$\pm$0.13 & 57.37$\pm$0.37 & \underline{68.21$\pm$0.19}& 66.32$\pm$0.25 & 66.49$\pm$0.35 & 65.36$\pm$0.31& \textbf{71.32$\pm$0.38} & $\uparrow$3.11 & 75.47$\pm$0.20 \\
    & 0.500\% & 60.70$\pm$0.15 & 58.67$\pm$0.28 & 59.11$\pm$0.39 & 66.38$\pm$0.49& \underline{67.41$\pm$0.58} & 66.45$\pm$0.39 & 65.32$\pm$0.31& \textbf{70.63$\pm$0.31} & $\uparrow$3.22& \\
    \midrule
    & 0.100\% & 51.06$\pm$0.09 & 52.98$\pm$0.18 & 52.44$\pm$0.23 & \underline{62.32$\pm$0.12}& 57.91$\pm$0.09 & 57.01$\pm$0.34 & 57.70$\pm$0.14& \textbf{63.84$\pm$0.12} & $\uparrow$1.52 & \\
    Amazon-Cloth & 0.300\% & 52.62$\pm$0.60 & 54.95$\pm$0.39 & 55.21$\pm$0.33 & \underline{62.49$\pm$0.06}& 60.96$\pm$0.21 & 56.94$\pm$0.37 & 59.85$\pm$0.08& \textbf{65.61$\pm$0.17} & $\uparrow$3.12 & 68.31$\pm$0.10 \\
    & 0.500\% & 54.38$\pm$0.53 & 55.78$\pm$0.34 & 56.86$\pm$0.35 & 62.00$\pm$0.03& 62.37$\pm$0.06 & 57.00$\pm$0.39 & \underline{62.89$\pm$0.33}& \textbf{65.26$\pm$0.14} & $\uparrow$2.37 & \\
    \bottomrule
    \end{tabular}
\end{table*}

\begin{figure}
    \centering
    \includegraphics[width=\columnwidth]{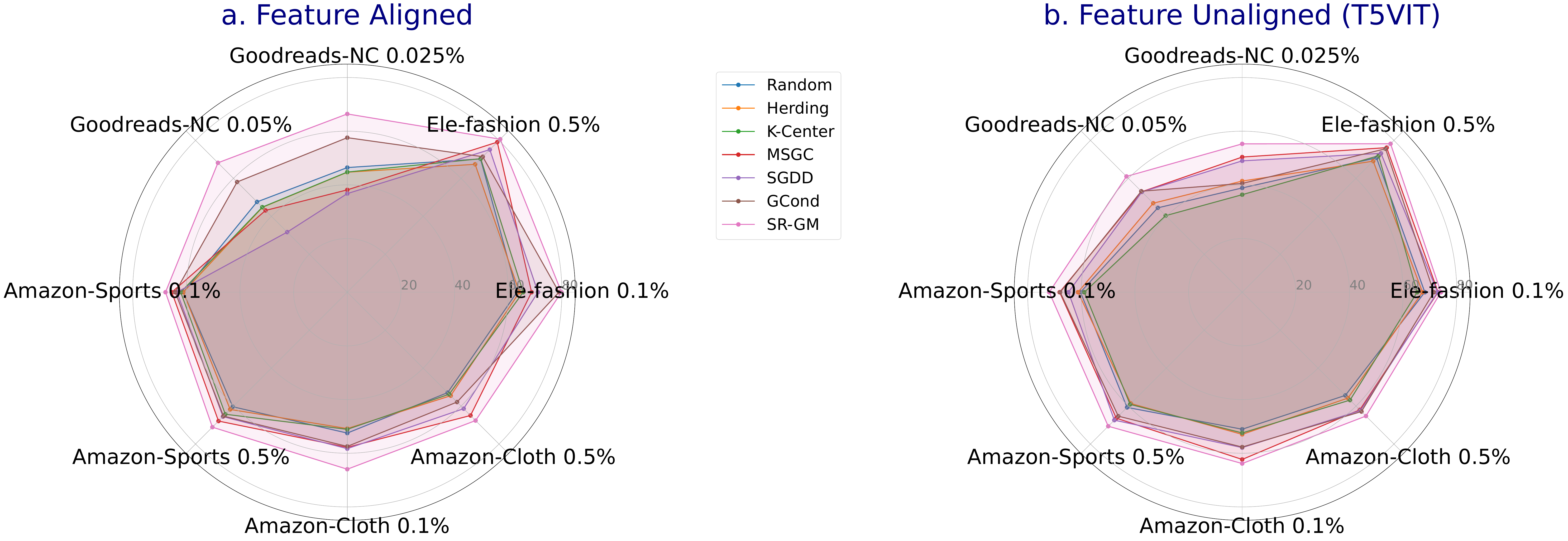}
    \caption{Average performance comparison between feature aligned and feature misaligned }
    \label{fig:comparison_both}
\end{figure}

\subsection{Analysis of Feature Alignment Effects on Multimodal Graph Condensation (Q2)}
We conduct multimodal graph condensation experiments with unaligned feature encoding to simulate real-world scenarios with feature inconsistency, where textual features are encoded using T5 \cite{raffel2020exploring} (512-dimensional) and visual features using ViT \cite{dosovitskiy2020image} (512-dimensional). This heterogeneous encoding creates semantic gaps in the feature space, providing a challenging environment to validate SR-GM's robustness. As shown in Table~\ref{tab:baseline_comp_t5vit}, SR-GM maintains superior performance across all datasets and condensation ratios under unaligned conditions, achieving improvements of 7.81\% over the sub-optimal method on Goodreads-NC at 0.05\% condensation ratio and 3.11\% on Amazon-Sports at 0.3\% condensation ratio, confirming the effectiveness of gradient decoupling in mitigating multimodal gradient conflicts.

Comparative analysis between Table~\ref{tab:baseline_comp} (aligned features) and Table~\ref{tab:baseline_comp_t5vit} (unaligned features) reveals that aligned conditions yield higher original dataset accuracy, with Ele-fashion achieving 85.68\% versus 84.47\%, Goodreads-NC reaching 79.45\% versus 76.93\%, and Amazon-Cloth attaining 71.59\% versus 68.31\%, reinforcing the crucial role of feature alignment in multimodal graph learning. However, the radar charts in Fig.~\ref{fig:comparison_both} demonstrate SR-GM's distinctive advantage: in both aligned and unaligned scenarios, it exhibits the smallest performance gap relative to original datasets, with Fig.~\ref{fig:comparison_both} particularly showing significantly smaller performance degradation compared to other condensation algorithms under unaligned conditions.

These findings highlight SR-GM's unique capability to maintain excellent performance in unaligned environments through its innovative gradient decoupling and structural regularization mechanisms, substantially reducing dependency on perfect feature alignment. This strong adaptability to unaligned feature environments positions SR-GM as a powerful solution for practical multimodal graph condensation challenges, especially considering that achieving perfect feature alignment in real-world applications is often prohibitively expensive or practically infeasible.

\begin{table}
    \centering
    \caption{Average training time per epoch (seconds) for GCond and SR-GM.}
    \label{tab:training_time}
    \small 
    \setlength{\tabcolsep}{8pt}
    \begin{tabular}{lccc}
    \toprule
    Dataset & Ratio ($r$) & GCond(s) & SR-GM(s) \\
    \midrule
            & 0.025\% & 8.36 & 9.40 \\
        \raisebox{1.5ex}[0pt]{Goodreads-NC}    & 0.050\% & 8.55 & 9.42 \\
    \midrule
            & 0.100\% & 2.11 & 2.01 \\
        Ele-fashion    & 0.300\% & 2.30 & 2.33 \\
            & 0.500\% & 2.42 & 2.73 \\
    \midrule
            & 0.100\% & 2.06 & 2.18 \\
        Amazon-Sports    & 0.300\% & 2.13 & 2.21 \\
            & 0.500\% & 2.02 & 2.27 \\
    \midrule
            & 0.100\% & 2.10 & 2.50 \\
        Amazon-Cloth    & 0.300\% & 2.21 & 2.69 \\
            & 0.500\% & 2.35 & 2.92 \\
    \bottomrule
    \end{tabular}
\end{table}

\begin{figure}
    \centering
    \includegraphics[width=\columnwidth]{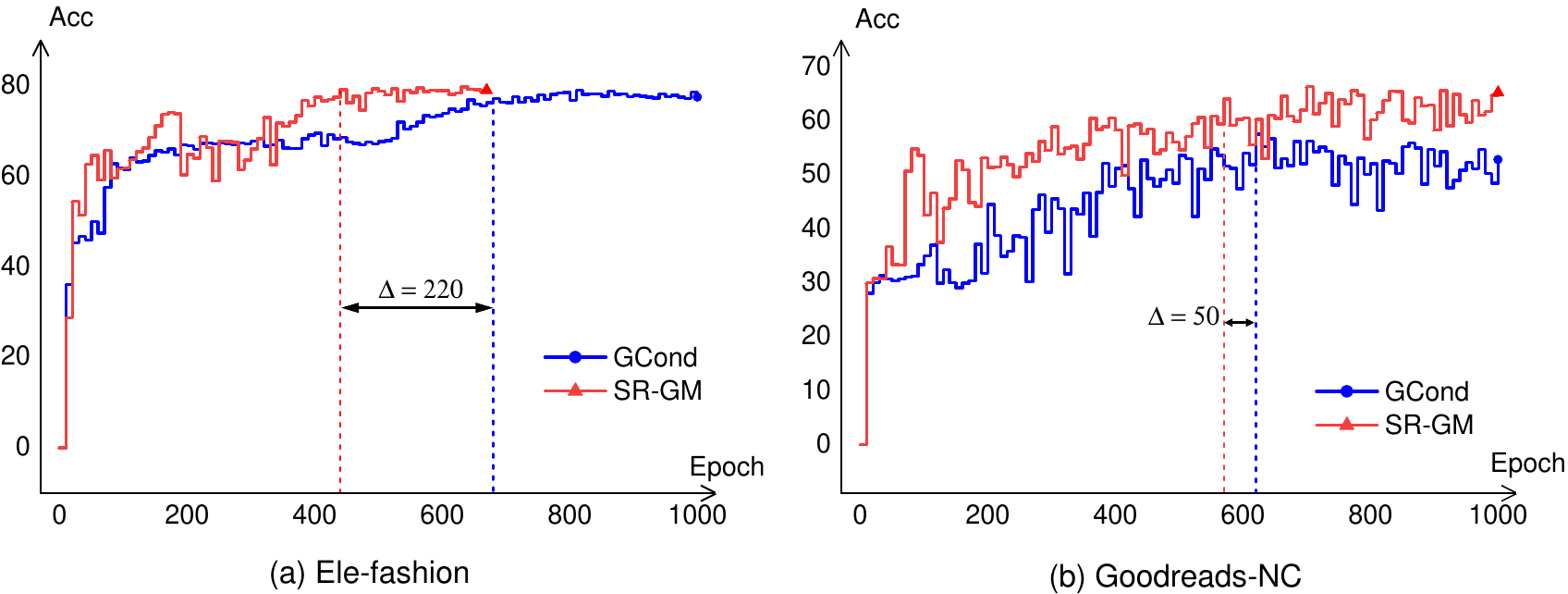}
    \caption{Comparison of condensation efficiency between our proposed SR-GM and the baseline GCond on the Ele-fashion and Goodreads-NC datasets. We can observe that SR-GM achieves peak performance with a faster convergence speed.}
    \label{fig:conden_efficiency}
\end{figure}

\begin{table*}
\caption{SR-GM Ablation Study Results. D: gradient decoupling, R: structural regularization. $\Delta_1$: (SR-GM) - (SR-GM w/o D), $\Delta_2$: (SR-GM w/o D) - (SR-GM w/o D+R).}
\label{tab:ablation_study_simple}
\centering
\small
\setlength{\tabcolsep}{4pt}
\begin{tabular}{lcccccc}
\toprule
Dataset & Ratio ($r$) & Method & CLIP (Aligned) & T5-ViT (Unaligned) & CLIP Increments & T5-ViT Increments \\
\midrule
\multirow{6}{*}{Goodreads-NC} 
 & \multirow{3}{*}{0.025\%} & SR-GM & 66.43$\pm$0.60 & 55.28$\pm$0.11 & - & - \\
 & & w/o D & 65.15$\pm$0.66 & 54.34$\pm$0.32 & $\Delta_1$: +1.28& $\Delta_1$: +0.94\\
 & & w/o D+R & 57.59$\pm$0.55 & 40.53$\pm$0.30 & $\Delta_2$: +7.56& $\Delta_2$: +13.81\\
\cmidrule{2-7}
 & \multirow{3}{*}{0.050\%} & SR-GM & 68.20$\pm$0.31 & 61.03$\pm$0.21 & - & - \\
 & & w/o D & 66.97$\pm$0.23 & 60.89$\pm$0.20 & $\Delta_1$: +1.23& $\Delta_1$: +0.14\\
 & & w/o D+R & 58.09$\pm$0.66 & 53.22$\pm$0.18 & $\Delta_2$: +8.88& $\Delta_2$: +7.67\\
\midrule
\multirow{9}{*}{Ele-fashion}
 & \multirow{3}{*}{0.100\%} & SR-GM & 79.85$\pm$0.22 & 74.09$\pm$0.13 & - & - \\
 & & w/o D & 79.63$\pm$0.27& 71.09$\pm$0.48 & $\Delta_1$: +0.22& $\Delta_1$: +3.00\\
 & & w/o D+R & 78.93$\pm$0.06 & 71.44$\pm$0.22 & $\Delta_2$: +0.92& $\Delta_2$: -0.35\\
\cmidrule{2-7}
 & \multirow{3}{*}{0.300\%} & SR-GM & 82.03$\pm$0.13 & 77.47$\pm$0.14 & - & - \\
 & & w/o D & 81.95$\pm$0.06 & 73.47$\pm$0.07 & $\Delta_1$: +0.08& $\Delta_1$: +4.00\\
 & & w/o D+R & 68.00$\pm$0.78 & 76.91$\pm$0.11 & $\Delta_2$: +13.95& $\Delta_2$: -3.44\\
\cmidrule{2-7}
 & \multirow{3}{*}{0.500\%} & SR-GM & 80.64$\pm$0.17 & 78.23$\pm$0.13 & - & - \\
 & & w/o D & 81.82$\pm$0.25 & 75.65$\pm$0.25 & $\Delta_1$: -1.18& $\Delta_1$: +2.58\\
 & & w/o D+R & 71.42$\pm$0.28 & 78.55$\pm$0.07 & $\Delta_2$: +10.40& $\Delta_2$: -2.90\\
\bottomrule
\end{tabular}
\end{table*}

\subsection{Condensation Efficiency (Q3)}
We evaluate the computational efficiency of SR-GM by comparing its convergence behavior and training latency against the GCond. Specifically, we analyze the convergence trajectory of test accuracy relative to training epochs and assess the potential overhead introduced by our method.

As illustrated in Fig.~\ref{fig:conden_efficiency}, SR-GM converges faster by 220 and 50 epochs on the Ele-fashion and Goodreads-NC datasets, respectively. And Table~\ref{tab:training_time} indicates that the average per-epoch training time of SR-GM remains nearly identical to that of GCond. These results demonstrate that SR-GM not only achieves superior performance but also maintains a competitive condensation speed.

\subsection{Ablation Study on Gradient Decoupling and Structural Regularization (Q4)} 
The ablation study results in Table~\ref{tab:ablation_study_simple} fully validate the effectiveness of the SR-GM framework in addressing gradient conflict issues in multimodal graph condensation. In feature-aligned CLIP encoding scenarios, structural regularization (R) demonstrates significant advantages, particularly on the Goodreads-NC dataset where the $\Delta_2$ increments reach +7.56 to +8.88. This confirms our theoretical hypothesis regarding ``structural damping'' - by constraining the gradient field through Laplacian regularization, it effectively suppresses the propagation and amplification of local gradient conflicts within the graph structure. 

However, in feature-unaligned T5-ViT encoding environments on the Ele-fashion dataset, the method relying solely on structural regularization ($\Delta_2 = -3.44$) shows limitations, reflecting that structural constraints alone may be insufficient to completely resolve the deep optimization problems caused by inter-modal semantic gaps.
In contrast, the core value of gradient decoupling (D) becomes prominent. The stable $\Delta_1$ increments of +2.58 to +4.00 on the Ele-fashion dataset directly confirm the crucial role of the gradient decoupling mechanism in handling conflicts among multimodal gradient signals. Particularly noteworthy is the negative $\Delta_2$ increment observed for structural regularization under T5-ViT encoding, a phenomenon that profoundly reveals how traditional structural constraints might instead reinforce incorrect gradient propagation paths in feature-unaligned environments. Gradient decoupling, through orthogonal projection to eliminate inter-modal gradient contradictions, provides a cleaner optimization foundation for structural regularization.

These findings strongly support the dual-design philosophy of SR-GM: gradient decoupling resolves multimodal gradient conflicts at their source, while structural regularization suppresses conflict amplification along propagation paths. The synergistic effect of these two components enables SR-GM to maintain robust performance across diverse multimodal graph scenarios, providing reliable technical support for practical applications of multimodal graph condensation.

\subsection{Ablation Study on the Impact of Multimodal Features (Q5)}
To investigate the impact of multimodal features on the performance of graph condensation, we test the performance of GCond when condensing three types of input features: text features, image features and fused text-image features. Then we compare it against SR-GM when condensing fused features. 

\begin{figure}
    \centering
    \includegraphics[width=0.9\columnwidth]{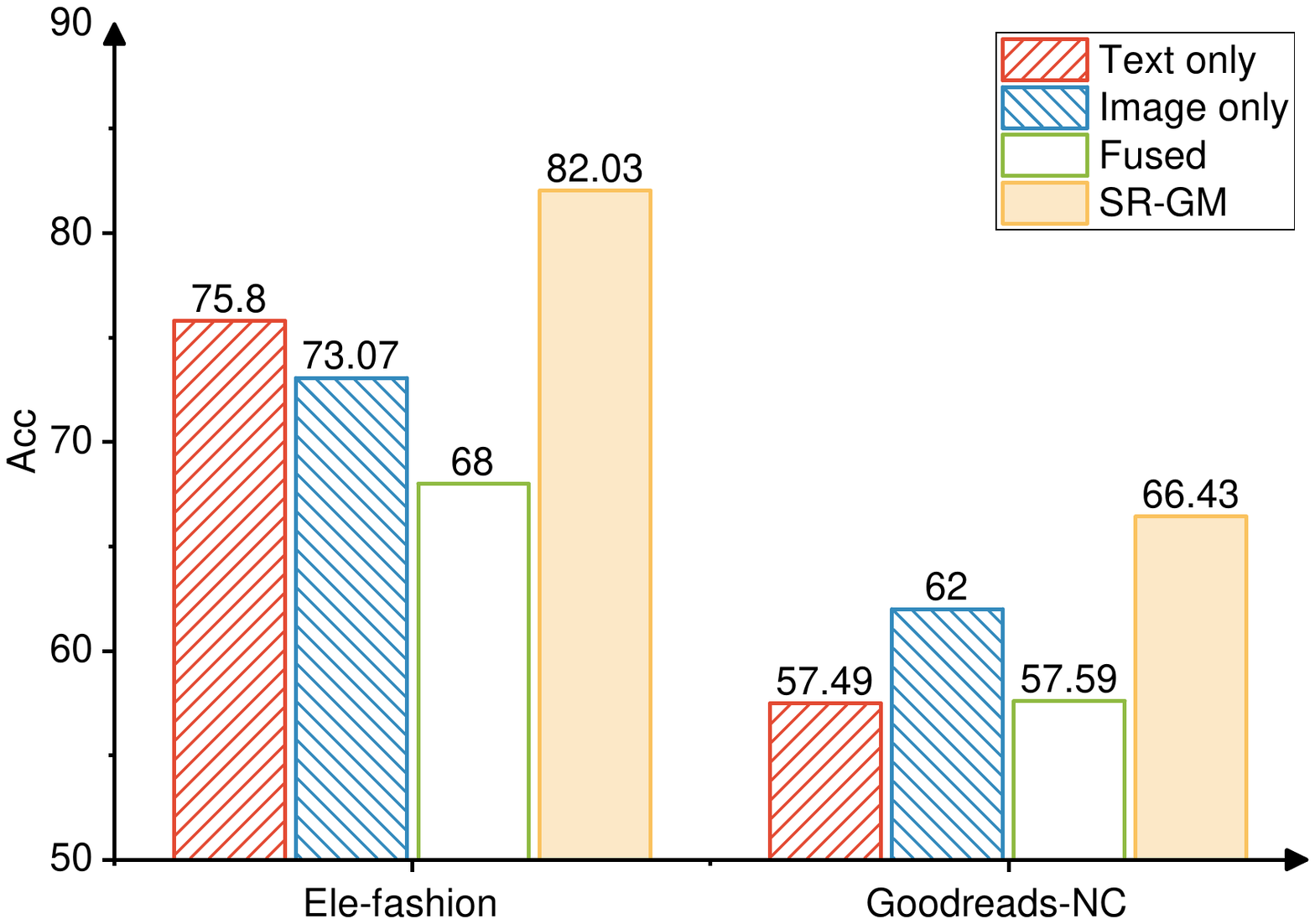}
    \caption{
        Test accuracy (\%) for SR-GM and GCond using different input features: text-only, image-only, and fused text-image features.
    }
    \label{fig:multimodal_impact}
\end{figure}

The node classification accuracy at Ele-fashion and Goodreads-NC is presented in Fig.~\ref{fig:multimodal_impact}. The fused multimodal data reduces the performance of GCond by approximately 6\%. This may due to the inherent conflict between different modalities. 
In contrast, SR-GM outperform GCond by 14\% at Ele-fashion and 9\% at Goodreads-NC. It shows SR-GM's ability to handle the multimodal graph data.

\section{Conclusion}
In this paper, we identify two fundamental challenges leading to the failure of current multimodal graph condensation methods: gradient conflicts arising from synthetic node features and the pathological amplification of gradient noise by graph structures. To address these issues, we propose a novel framework, Structurally-Regularized Gradient Matching (SR-GM), for condensing multimodal graph data. Specifically, SR-GM tackles the problem through a bi-level optimization formulation that incorporates two synergistic components. First, a gradient disentanglement mechanism resolves inter-modal conflicts at their source via orthogonal projection. Second, a structural damping regularizer, based on the Dirichlet energy of the gradient field, transforms the graph topology from a noise amplifier into an optimization stabilizer. Experimental results demonstrate the superiority of the proposed SR-GM in both efficiency and effectiveness. Furthermore, SR-GM exhibits strong generalization capabilities across various GNN architectures on multimodal graph-structured data.

\bibliographystyle{ACM-Reference-Format}
\bibliography{SR-GM}

\appendix

\section{APPENDIX}
\label{appendix}

\subsection{Related Work}
\subsubsection{Graph Dataset Condensation.} Dataset Condensation (DC), also known as Dataset Distillation, aims to synthesize a small, informative dataset from a large original one such that models trained on the synthetic set achieve performance comparable to those trained on the original data. This paradigm was first pioneered in the image domain, where methods were developed to optimize synthetic pixels directly through a bi-level optimization framework \cite{wang2018dataset}. The core idea is to match certain properties between the real and synthetic data. Various matching criteria have been proposed, including matching model performance \cite{zhao2020dataset} and feature distributions \cite{zhao2023dataset,wang2022cafe}. Among these, matching the gradients of model parameters has been proven to be a particularly effective strategy \cite{zhao2020dataset}. The extension of these techniques to the graph domain, termed Graph Condensation (GC), is a nascent and challenging endeavor. Jin et al. \cite{jin2021graph} first extended DC to graphs, introducing a node-level condensation method. Unlike image data, graph data involves an intricate coupling between node attributes and topological structure, making the joint synthesis of features and edges a complex optimization task. Current GC methods primarily focus on preserving information in single-modal graphs, exploring strategies such as gradient matching \cite{jin2021graph}, training trajectory matching \cite{zheng2023structure}, and distribution matching \cite{liu2022graph}. Recent research has evolved from developing efficient and interpretable graph condensation methods \cite{fang2024exgc,wang2024fast} to creating specialized techniques tailored for specific graph types, such as heterogeneous graphs and signed graphs \cite{gao2024heterogeneous,li2025structure}. However, these methods largely overlook the unique challenges posed by multimodal graphs, where node features themselves are composed of heterogeneous information from sources like text and images. The problem of inter-modal signal conflict within the GC framework remains a critical and underexplored gap. Our work aims to bridge this gap.

\subsubsection{Multimodal Graph Learning.} Multimodal Graph Learning (MGL) has emerged as a vital field for analyzing complex real-world systems where entities are described by multiple data types (e.g., text, images, audio). MGL aims to leverage the relational structure of graphs to fully explore associations both within and between different modalities. This is particularly challenging because data fusion must be guided by the graph topology, which itself may vary across modalities. A key challenge in MGL is bridging the "semantic gap", where different modalities may provide conflicting or inconsistent representations of the same entity \cite{zhu2025mosaic,liang2022mind}. For instance, two products might be visually similar but have different textual descriptions, creating discrepancies that unimodal models cannot resolve. To address this, significant research has focused on feature alignment, which aims to map representations from different modalities into a unified embedding space. Contrastive learning frameworks, such as CLIP \cite{radford2021learning}, have been notably successful in learning aligned vision-language representations. From an architectural perspective, methods like Multimodal Graph Convolutional Networks (MGCN) \cite{wei2019mmgcn} and Multimodal Graph Attention Networks (MGAT) \cite{tao2020mgat} have been proposed to explicitly model and fuse information from different modalities. While these MGL methods focus on designing more sophisticated Graph Neural Network (GNN) architectures to better handle existing multimodal data, they do not address the data-centric challenge of scalability \cite{ninggraph4mm,cai2024multimodal}. The question of how to effectively compress large-scale multimodal graphs, particularly in a way that handles the feature conflicts inherent in the condensation process, remains largely unaddressed. Our work bridges this gap by introducing a condensation method specifically designed with an optimization scheme attuned to the uniqueness of multimodal graphs.

\subsection{Complexity Analysis}
We analyze the computational complexity of the proposed SR-GM framework, focusing on the additional overhead introduced by the gradient decoupling and structural regularization modules. 

\subsubsection{Gradient Decoupling Complexity}
The gradient decoupling process operates primarily on the gradient vectors of the synthetic nodes. As shown in the Algorithm~\ref{alg:sr-gm}, it involves computing element-wise dot products, norms, and vector projections for the text and image modalities. These operations are linear with respect to the number of synthetic nodes and do not involve node-to-node interactions. 
For $N'$ synthetic nodes with $d$-dimensional features, the complexity of this component is $\mathcal{O}(N' d)$.

\subsubsection{Structural Regularization Complexity}
The structural regularization minimizes the Dirichlet energy of the decoupled gradient field, formulated as Eq.~(\ref{eq:structure_match_loss}). 
The computational bottleneck lies in the matrix multiplication, which has a complexity of $\mathcal{O}((N')^2 d)$. The subsequent trace operation takes $\mathcal{O}(N' d)$. Thus, the complexity is dominated by $\mathcal{O}((N')^2 d)$. 

\subsubsection{Overall Time Complexity}
Combining the two components, the total additional complexity is dominated by $\mathcal{O}((N')^2 d)$.
It is important to note that the size of the condensed graph $N'$ is extremely small. Consequently, the term $(N')^2$ remains negligible compared to the complexity of training GNNs on the original large-scale graph. Therefore, SR-GM improves condensation quality with marginal computational overhead.

\subsection{Generalizability Analysis (Q6)}
To demonstrate that the effectiveness of our proposed SR module is not confined to a condensation paradigm or neural architecture, we investigate the generalizability of SR-GM from three perspectives: (i) compatibility with different matching objectives, (ii) transferability of the condensed graphs across various GNN backbones, and (iii) the performance stability when employing different architectures during the condensation process.

\begin{table}
    \caption{Performance and efficiency comparison between GCDM and SR-GCDM. All datasets use CLIP encoding. Performance is reported as test accuracy (\%) for Ele-fashion, Goodreads-NC and AUC score for Amazon-Sports, Amazon-Cloth. Total time required for condensation to converge is measured in seconds. The better accuracy is bolded.}
    \label{tab:dm_generalization}
    \centering
    \small
    \setlength{\tabcolsep}{2pt}
    \begin{tabular}{lccccc} 
    \toprule
    \multicolumn{2}{c}{} & \multicolumn{2}{c}{Accuracy} & \multicolumn{2}{c}{Time(s)} \\
    \cmidrule(lr){3-4} \cmidrule(lr){5-6}
    \raisebox{1.5ex}[0pt]{Dataset} & \raisebox{1.5ex}[0pt]{Ratio ($r$)} & GCDM & SR-GCDM & GCDM & SR-GCDM  \\
    \midrule
    & 0.100\% & 76.60$\pm$0.09 & \textbf{79.06$\pm$0.10} & 65.05 & 83.76 \\
    Ele-fashion & 0.300\% & 77.21$\pm$0.06 & \textbf{79.76$\pm$0.05} & 95.59 & 85.46 \\
    & 0.500\% & 77.19$\pm$0.12 & \textbf{79.26$\pm$0.11} & 153.29 & 149.35 \\
    \midrule
    & 0.025\% & 42.08$\pm$0.32 & \textbf{51.19$\pm$0.52} & 453.53 & 525.12 \\
    \raisebox{1.5ex}[0pt]{Goodreads-NC} & 0.050\% & 43.56$\pm$0.43 & \textbf{56.08$\pm$0.47} & 284.86 & 481.75 \\
    \midrule
    & 0.100\% & 57.96$\pm$0.02 & \textbf{58.02$\pm$0.02} & 27.38 & 32.45 \\
    Amazon-Cloth & 0.300\% & \textbf{58.07$\pm$0.03} & 58.03$\pm$0.02 & 60.04 & 65.84 \\
    & 0.500\% & \textbf{58.40$\pm$0.03} & 58.28$\pm$0.06 & 55.52 & 59.51 \\
    \midrule
    & 0.100\% & 65.46$\pm$0.34 & \textbf{66.94$\pm$0.09} & 5.71 & 12.49 \\
    Amazon-Sports & 0.300\% & 65.59$\pm$0.31 & \textbf{66.69$\pm$0.08} & 55.93 & 47.52 \\
    & 0.500\% & 65.42$\pm$0.38 & \textbf{65.62$\pm$0.18} & 63.88 & 68.78 \\
    \bottomrule
    \end{tabular}
\end{table}

\subsubsection{Generalization to Distribution Matching}
To evaluate the versatility of the SR module, we extend its application to Distribution Matching (DM). Specifically, we integrate the SR module into \textbf{GCDM}, denoted as \textbf{SR-GCDM}. 

Table~\ref{tab:dm_generalization} summarizes the performance and efficiency comparison between the original GCDM and our SR-enhanced version. Specifically, SR improves GCDM's performance on most datasets, with gains of approximately $3\%$ on Ele-fashion and $10\%$ on Goodreads-NC. Similar to our observations with SR-GM, while the incorporation of the SR module introduces a slight computational overhead, the increase in total time required for condensation to converge is marginal compared to the performance boost. This demonstrates that SR is a plug-and-play component that is agnostic to the specific matching objective.

\begin{table}
    \centering
    \small
    \setlength{\tabcolsep}{3pt}
    \caption{Evaluation of cross-architecture generalizability. Test accuracy (\%) of various GNN architectures trained on a graph condensed by SR-GM (with a GCN backbone).SAGE stands for GraphSAGE. Avg. stands for the average test accuracy.}
    \label{tab:generalizability}
    \begin{tabular}{cccccccc}
    \toprule
    Dataset & Methods & GCN & MLP & MMGCN & GAT & SAGE & Avg. \\
    \midrule
    Ele-fashion & GCond & 78.87 & 74.94 & 71.53 & 34.65 & 64.46 & 64.89 \\
    $r=0.1\%$ & SR-GM & 79.91 & 81.71 & 75.88 & 39.31 & 81.50 & 71.66 \\
    \midrule
    Goodreads-NC & GCond & 62.07 & 49.06 & 43.81 & 33.41 & 53.56 & 48.38 \\
    $r=0.05\%$ & SR-GM & 67.76 & 57.34 & 51.36 & 34.33 & 64.16 & 54.99 \\
    \bottomrule
    \end{tabular}
\end{table}

\subsubsection{Cross-Architecture Transferability}
Condensed graphs should exhibit transferability across diverse GNN architectures, regardless of the model used during the condensation. We condense the datasets using SR-GM with a GCN in the condensation process and subsequently evaluate the performance of the condensed graphs on several unseen architectures.

As shown in Table~\ref{tab:generalizability}, we test GraphSAGE \cite{hamilton2017inductive}, GAT \cite{velickovic2017graph}, a multimodal GNN (MMGCN) \cite{wei2019mmgcn}, and a simple MLP \cite{taud2017multilayer}. And we choose Ele-fashion and Goodreads-NC to report the performance. The graphs condensed by SR-GM show strong transfer performance on other architectures, maintaining high accuracy when the downstream model differs from the condensation model.

\begin{table}
    \centering
    \caption{Test accuracy (\%) of SR-GM on the Ele-fashion datasets when using different backbone models for both condensation and evaluation. "SAGE" denotes GraphSAGE. "Original" means training on original graph. The best results are highlighted in bold.}
    \label{tab:arch_agnostic_results}
    \begin{tabular}{lcccc}
    \toprule
    Backbone & Ratio ($r$) & SR-GM & GCond & Original \\
    \midrule
    & 0.1\% & \textbf{79.85$\pm$0.22} & 78.93$\pm$0.06 &  \\
    \raisebox{1.5ex}[0pt]{GCN} & 0.5\% & \textbf{80.64$\pm$0.17} & 71.42$\pm$0.28 & \raisebox{1.5ex}[0pt]{85.68$\pm$0.08}  \\
    \midrule
    & 0.1\% & \textbf{81.20$\pm$0.09} & 80.17$\pm$0.58 &  \\
    \raisebox{1.5ex}[0pt]{SAGE} & 0.5\% & \textbf{82.66$\pm$0.21} & 82.08$\pm$0.24 & \raisebox{1.5ex}[0pt]{87.52$\pm$0.01}  \\
    \midrule
    & 0.1\% & \textbf{77.27$\pm$0.34} & 77.07$\pm$0.51 &  \\
    \raisebox{1.5ex}[0pt]{MLP} & 0.5\% & \textbf{78.95$\pm$0.38} & 78.28$\pm$0.52 & \raisebox{1.5ex}[0pt]{88.37$\pm$0.03}  \\
    \bottomrule
    \end{tabular}
\end{table}

\subsubsection{Architectural Agnosticism of the Condensation Process}
We investigate whether the condensation performance of SR-GM is sensitive to the choice of the GNN model used during the matching process. As shown in Table~\ref{tab:arch_agnostic_results}, we report the performance at Ele-fashion. For each run, the same model (e.g., GCN) is used for condensation and downstream evaluation. 

From the table, we observe that SR-GM consistently constructs effective condensed graphs across all tested architectures. This validates that SR-GM does not need a specific GNN architecture to achieve superior condensation results.

\begin{figure}
    \centering
    \includegraphics[width=0.9\columnwidth]{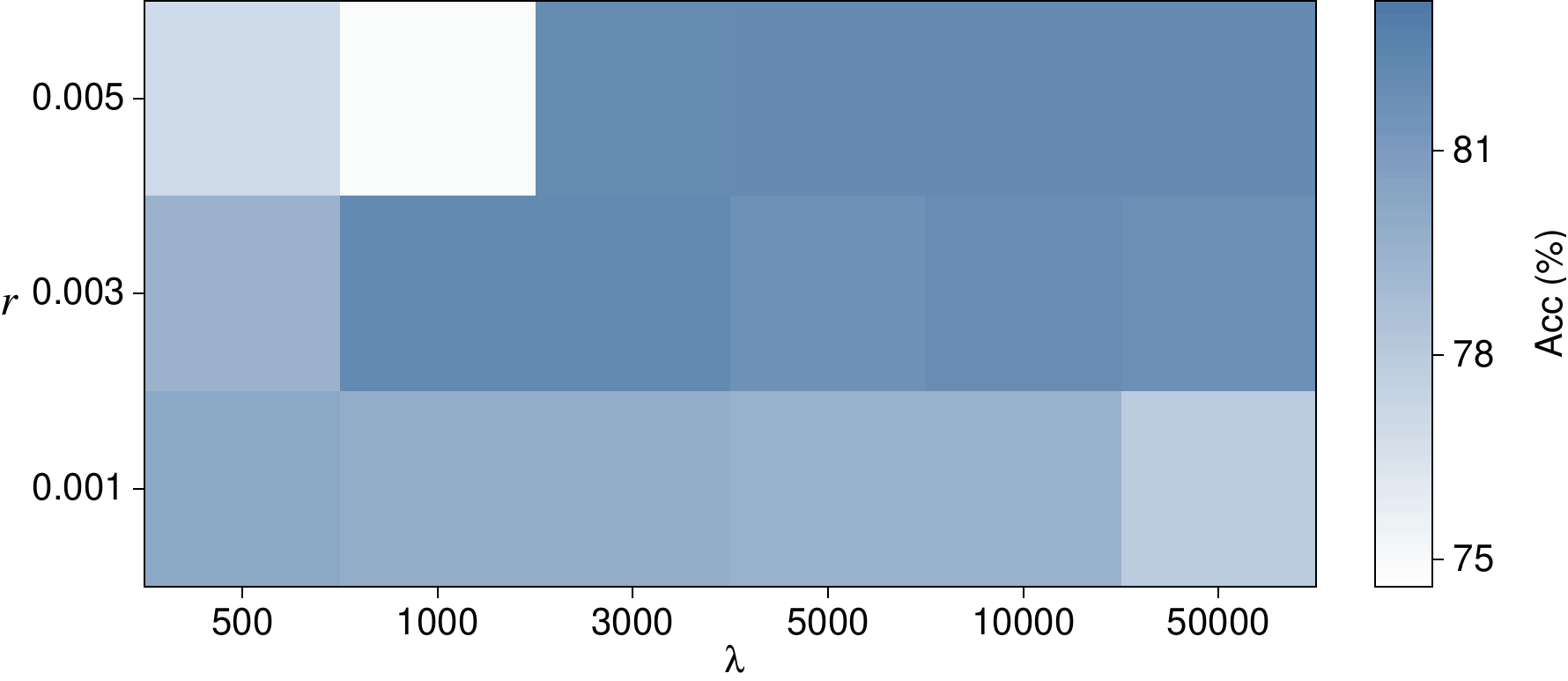}
    \caption{Heatmap of test accuracy for SR-GM on the Ele-fashion dataset under different settings of condensation rate ($r$) and regularization weight ($\lambda$). Darker shades correspond to higher accuracy, indicating better performance.}
    \label{fig:heatmap_lambda_r}
\end{figure}

\subsection{Analysis of Regularization Weight \texorpdfstring{$\lambda$}{lambda} (Q6)}
A critical hyperparameter in SR-GM is the regularization weight, denoted by $\lambda$. To investigate its impact on the graph condensation process, we condense graphs with various settings of $\lambda$ and condensation ratio, then report the accuracies in Fig.~\ref{fig:heatmap_lambda_r}. The darker shades indicate higher accuracy. 

From the heatmap, we observe that SR-GM has robust performance across most configurations. A notable exception occurs at $r$ of 0.05\%, where performance degrades for $\lambda$ values of 500 and 1000. It shows that while SR-GM is insensitive to high values of $\lambda$, a correlation still exists between $r$ and $\lambda$. Specifically, we should use a larger $\lambda$ when $r$ increases to maintain optimal performance.

\subsection{Proof of Theorem \ref{thm:noise_amplification} }\label{proof_t}

\begin{restate}{thm:noise_amplification}
Under gradient matching optimization, the amplification factor of modal mixing noise propagated through the graph structure is bounded by the Dirichlet energy of the gradient field:
\[
\|\mathbf{L}_S \mathbf{R}\|_F^2 \leq \lambda_{\max}(\mathbf{L}_S) \cdot E(\mathbf{R}),
\]
where \(E(\mathbf{R}) = \operatorname{tr}(\mathbf{R}^\top \mathbf{L}_S \mathbf{R})\) is the Dirichlet energy of the gradient field, and \(\lambda_{\max}(\mathbf{L}_S)\) is the maximum eigenvalue of the Laplacian matrix.
\end{restate}

\begin{proof}
We first define the gradient noise matrix \(\mathbf{R} \in \mathbb{R}^{N' \times d}\), where each row \(\mathbf{r}_i^\top\) represents the gradient noise of node \(i\). The graph Laplacian matrix \(\mathbf{L}_S \in \mathbb{R}^{N' \times N'}\) is symmetric positive semi-definite and can be decomposed via eigenvalue decomposition:
\[
\mathbf{L}_S = \mathbf{U} \boldsymbol{\Lambda} \mathbf{U}^\top,
\]
where \(\mathbf{U}\) is an orthogonal matrix, \(\boldsymbol{\Lambda} = \operatorname{diag}(\lambda_1, \lambda_2, \dots, \lambda_{N'})\), and the eigenvalues satisfy \(0 = \lambda_1 \leq \lambda_2 \leq \cdots \leq \lambda_{N'} = \lambda_{\max}(\mathbf{L}_S)\).

After propagation through the Laplacian matrix, the noise becomes \(\mathbf{L}_S \mathbf{R}\), and the square of its Frobenius norm is:
\[
\|\mathbf{L}_S \mathbf{R}\|_F^2 = \operatorname{tr}\left( \mathbf{R}^\top \mathbf{L}_S^\top \mathbf{L}_S \mathbf{R} \right) = \operatorname{tr}\left( \mathbf{R}^\top \mathbf{L}_S^2 \mathbf{R} \right).
\]

Substituting the eigenvalue decomposition:
\[
\|\mathbf{L}_S \mathbf{R}\|_F^2 = \operatorname{tr}\left( \mathbf{R}^\top \mathbf{U} \boldsymbol{\Lambda}^2 \mathbf{U}^\top \mathbf{R} \right) = \operatorname{tr}\left( \tilde{\mathbf{R}}^\top \boldsymbol{\Lambda}^2 \tilde{\mathbf{R}} \right),
\]
where \(\tilde{\mathbf{R}} = \mathbf{U}^\top \mathbf{R}\). Let \(\tilde{\mathbf{r}}_i^\top\) be the \(i\)-th row of \(\tilde{\mathbf{R}}\), then:
\[
\|\mathbf{L}_S \mathbf{R}\|_F^2 = \sum_{i=1}^{N'} \lambda_i^2 \|\tilde{\mathbf{r}}_i\|^2.
\]

On the other hand, the Dirichlet energy is defined as:
\[
E(\mathbf{R}) = \operatorname{tr}\left( \mathbf{R}^\top \mathbf{L}_S \mathbf{R} \right) = \operatorname{tr}\left( \mathbf{R}^\top \mathbf{U} \boldsymbol{\Lambda} \mathbf{U}^\top \mathbf{R} \right) = \operatorname{tr}\left( \tilde{\mathbf{R}}^\top \boldsymbol{\Lambda} \tilde{\mathbf{R}} \right) = \sum_{i=1}^{N'} \lambda_i \|\tilde{\mathbf{r}}_i\|^2.
\]

Since all eigenvalues are non-negative and satisfy \(\lambda_i \leq \lambda_{\max}\), we have \(\lambda_i^2 \leq \lambda_{\max} \lambda_i\), thus:
\[
\sum_{i=1}^{N'} \lambda_i^2 \|\tilde{\mathbf{r}}_i\|^2 \leq \lambda_{\max} \sum_{i=1}^{N'} \lambda_i \|\tilde{\mathbf{r}}_i\|^2,
\]
which is equivalent to:
\[
\|\mathbf{L}_S \mathbf{R}\|_F^2 \leq \lambda_{\max}(\mathbf{L}_S) \cdot E(\mathbf{R}).
\] 

Equality holds if and only if all non-zero components of \(\mathbf{R}\) lie in the eigenspace corresponding to the maximum eigenvalue. This inequality demonstrates that the amplification of gradient noise by the graph structure is bounded by the maximum eigenvalue of the Laplacian matrix and the Dirichlet energy of the gradient field. 

\end{proof}

\end{document}